\documentclass[lettersize,journal]{IEEEtran}
\usepackage{amsmath,amsfonts}
\usepackage{algorithmic}
\usepackage{algorithm}
\usepackage{array}
\usepackage[caption=false,font=normalsize,labelfont=sf,textfont=sf]{subfig}
\usepackage{textcomp}
\usepackage{stfloats}
\usepackage{url}
\usepackage{verbatim}
\usepackage{graphicx}
\usepackage{cite}
\usepackage{multirow}
\usepackage{booktabs}
\usepackage{multirow}
\begin{document}

\title{TC-SfM: Robust Track-Community-Based Structure-from-Motion}

\author{Lei Wang, Linlin Ge, Shan Luo, Zihan Yan, Zhaopeng Cui and Jieqing Feng
\thanks{This work was jointly supported by the National Natural Science Foundation of China under Grant Nos. 61732015 and 61932018. (\textit{Corresponding author: Jieqing Feng.})
}
\thanks{
	Lei Wang, Linlin Ge, Shan Luo, Zihan Yan, Zhaopeng Cui and Jieqing Feng are with the State Key Laboratory of CAD\&CG, Zhejiang University, Hangzhou 310058, China (e-mail: \{iwlei, linlinge, luoshan, zihanyan, zhpcui\}@zju.edu.cn; jqfeng@cad.zju.edu.cn).
}
}

\markboth{}%
{Shell \MakeLowercase{\textit{et al.}}: A Sample Article Using IEEEtran.cls for IEEE Journals}


\maketitle

\begin{abstract}
Structure-from-Motion (SfM) aims to recover 3D scene structures and camera poses based on the correspondences between input images, and thus the ambiguity caused by duplicate structures (i.e., different structures with strong visual resemblance) always results in incorrect camera poses and 3D structures. To deal with the ambiguity, most existing studies resort to additional constraint information or implicit inference by analyzing two-view geometries or feature points. In this paper, we propose to exploit high-level information in the scene, i.e., the spatial contextual information of local regions, to guide the reconstruction. Specifically, a novel structure is proposed, namely, {\textit{track-community}}, in which each community consists of a group of tracks and represents a local segment in the scene. A community detection algorithm is used to partition the scene into several segments. Then, the potential ambiguous segments are detected by analyzing the neighborhood of tracks and corrected by checking the pose consistency. Finally, we perform partial reconstruction on each segment and align them with a novel bidirectional consistency cost function which considers both 3D-3D correspondences and pairwise relative camera poses. Experimental results demonstrate that our approach can robustly alleviate reconstruction failure resulting from visually indistinguishable structures and accurately merge the partial reconstructions.  
\end{abstract}


\section{Introduction}
\IEEEPARstart{S}{tructure-from-Motion} is designed to recover camera motions and sparse 3D structures from image collections \cite{agarwal2011building, snavely2008modeling}. This technique has been applied to various scenarios, such as indoor-outdoor 3D reconstructions \cite{cui2022vidsfm, Zhang2016Eff}, natural environment monitoring\cite{clapuyt2016reproducibility}, cultural heritage digitization\cite{fuhrmann2014mve}, and recent neural rendering \cite{mildenhall2020nerf, wang2021neus}. The typical steps of SfM consist of feature detection, feature matching, camera poses estimation, and 3D structure reconstruction\cite{schonberger2016structure}.

Although SfM methods have achieved impressive performance across numerous tasks, 
the existing methods still struggle to reconstruct the scene with duplicate structures accurately, which are common in the real world, such as the repetitive facades and decorations in buildings. The reasons lie in the image feature matching. If different instances share a highly similar appearance, their local features tend to be falsely matched, which leads to the incorrect pose estimation as well as the final 3D reconstructions like superimposed and phantom structures.

Existing work often deals with ambiguity by analyzing the feature points or epipolar geometries (EGs) between two views to explore consistent constraints. For example, some work attempted to remove inconsistent EGs because the EGs contain more information, which makes it easier to be distinguished than feature points. To remove the incorrect EGs, they add additional geometric consistency constraints between the views, such as loop constraints \cite{zach2010disambiguating}. However, the effectiveness of these approaches is limited due to the accumulated geometric errors. Considering that the incorrect geometric relations stem from mismatched correspondences, a more fundamental solution is to analyze the visibility of points based on the correspondences, such as missing correspondences \cite{zach2008can} and visibility graph \cite{wilson2013network}. The co-occurrence of feature correspondences can provide additional inference about ambiguous structures. Nevertheless, these methods are based on the local triplets or each feature point individually, and are prone to remove many positive correspondences. Few of the previous methods exploit explicitly high-level information (i.e., scene structures), which is naturally exploited in human perception.

In this work, we exploit the underlying spatial contextual information of the local region of the scene, which provides additional spatial relationships, by grouping related points into the same cluster. Unlike the prior work that considers each point equally and ignores the underlying scene structures, our method considers their associated relationship. In this case, the ambiguous structure, which usually belongs to a local region of an object, can be directly detected at the region-level.

\begin{figure*}[!t]
	\centering
	\includegraphics[width=7in]{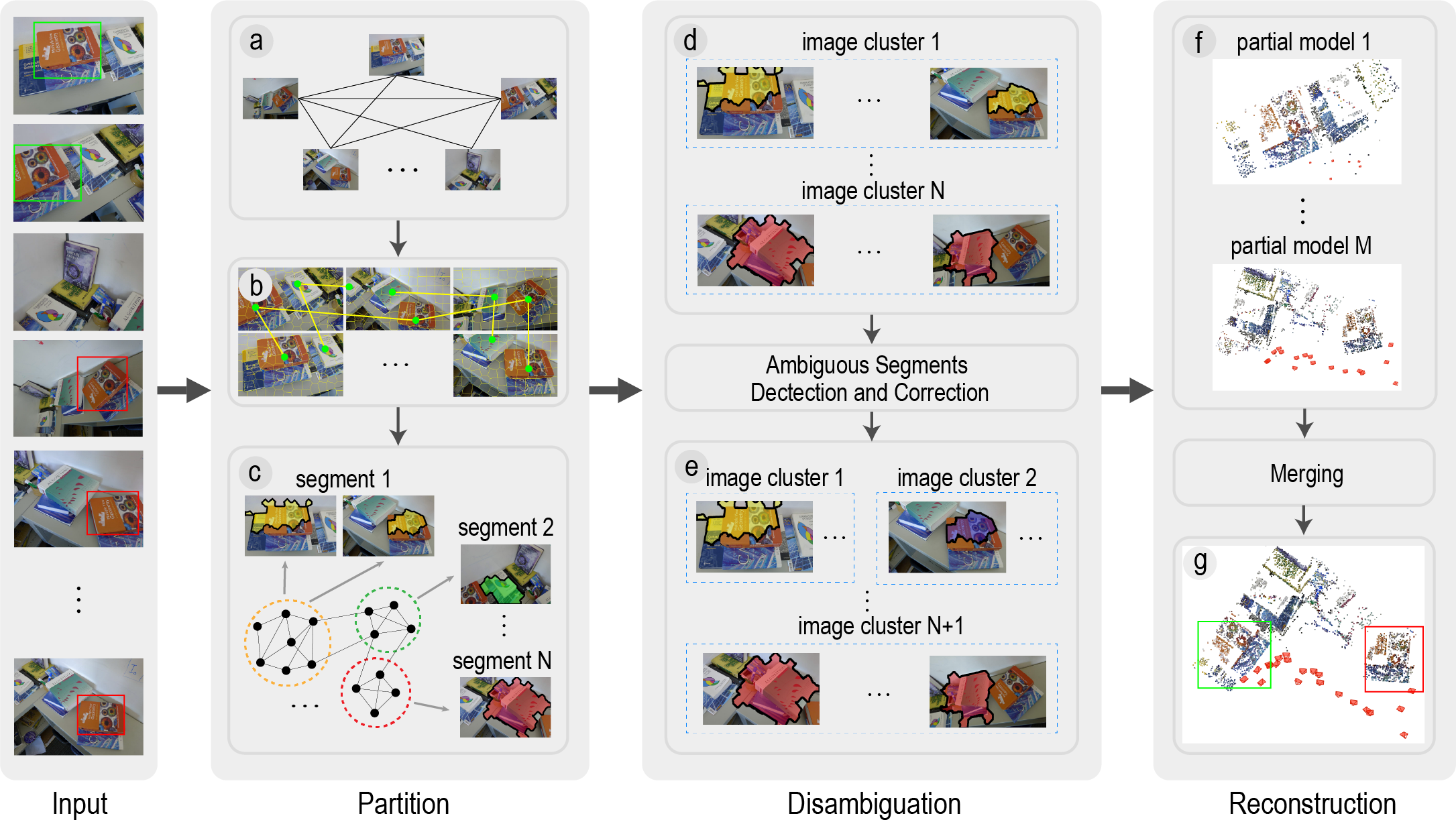}
	\caption{Pipeline  of the TC-SfM. Our method takes an image collection as input, and adopt the partition strategy for disambiguation and reconstruction. (a) View-graph construction. (b) Track sampling in superpixels. (c) Track-graph construction and community detection. (d) Image clusters corresponding the segments. (e) Image clusters after disambiguation. (f) 3D models of partial reconstructions. (g) Merged 3D models in a uniform frame.}
	\label{fig_1}
\end{figure*}

To this end, we propose a novel track-community structure to partition the scene into several parts without reconstructing the scene in advance. This structure is obtained by analyzing the adjacency of the tracks and performing a community detection algorithm. Specifically, a track is defined as a set of matched 2D feature points from different views and corresponds to a 3D point in the real world. Accordingly, tracks can encode the visibility of the 3D points in each view, and a track-community refers to the local region of an object or several adjacent objects in the scene, namely, a segment of the scene, as shown in Fig. \ref{fig_1}(c). Once the track-community is established, we can remove ambiguous structures via two steps, i.e., ambiguity detection and correction. In ambiguity detection, the diversity analysis of the track-graph is used to determine whether each track-community potentially contains erroneous tracks caused by ambiguous structures. In ambiguity correction, the pose confliction checking between segments is used to remove erroneous correspondences from ambiguous segments and recover correct poses.

For camera pose and 3D structure estimation, an incremental SfM approach is selected in our method because of its accuracy and robustness \cite{schonberger2016structure}. However, this approach usually suffers from the drift problem in large-scale reconstruction \cite{zhu2018very, Cui2015Large}. To overcome this drawback, many SfM methods adopt the strategy of first distributedly registering the cameras and then merging them \cite{zhu2018very, chen2020graph}. Considering that the whole scene has been divided into several parts in the previous disambiguation step, we also utilize the partitioning strategy for SfM. To refine the similarity transformation between partial 3D models, we propose a new merging algorithm to register all local reconstructions into a global frame, which takes both 3D-3D correspondences and pairwise EGs into account. 

In summary, the main contributions of this paper are listed as follows:
\begin{itemize}
	\item{We present a robust SfM method, i.e., TC-SfM, which explores the scene contextual information from track-communities to mitigate the problem of reconstruction failure caused by ambiguous structures.}
	\item{We propose a novel approach for detecting and correcting ambiguous structures by dividing the scene into several segments and checking the pose consistency among segments. }
	\item{A new merging algorithm using a bidirectional consistency cost function is proposed to accurately register all partial reconstructions into a unified framework.}
\end{itemize} 

We conduct experiments on various datasets, and the results show that TC-SfM can robustly alleviate reconstruction failure resulting from ambiguity and achieve superior performance.

\section{Related Work}
Many existing studies devote to improving the performance of SfM in the presence of ambiguous structures. These methods can be divided into three categories: optimizing view-graph construction, improving camera registration, and post-processing reconstruction results.

\subsubsection*{\bf View-graph construction} The view-graph, where the nodes are images from different views and the edges are pairwise EGs between them, is an indispensable component in many SfM pipelines. The reconstruction performance will be greatly influenced by the contaminative view-graph.

One solution to the view-graph optimization is to directly remove the incorrect edges in the view-graph. Zach et al. \cite{zach2008can} assumed that all three images of a triplet that share sufficient and well-distributed correspondences suggest the correct EG among all view pairs. The absence of enough correspondences (i.e., missing correspondences) provides strong evidence of the presence of an erroneous EG in them. Zach et al. \cite{zach2010disambiguating} enforced the loop consistency of geometric relations estimated from the input. They detected conflicting relations in a Bayesian framework to infer the set of likely false-positive geometric relations. Roberts et al.\cite{roberts2011structure} identified mismatched pairs based on an expectation-maximization framework that incorporates image timestamp cues with missing correspondence cues. However, the image timestamp information in an unordered dataset is difficult to obtain, thereby limiting its usage.  Wilson et al. \cite{wilson2013network} assumed that two observations visible on one view are also visible on other views and utilized the bipartite local clustering coefficient over the visibility graph to measure such consistency. This approach easily results in over-segmentation because the tracks with a score less than the threshold are all deleted.

Another solution is to optimize the edges, which aims to improve the quality of two-view geometry. Cui et al.\cite{cui2015global} performed a local bundle adjustment (BA) in pairwise images to improve relative motion. Sweeney et al. \cite{sweeney2015optimizing} improved the quality of the relative geometries in the triplet by enforcing loop consistency constraints with an epipolar point transfer. Other studies concentrate on finding an optimal subgraph from the full view-graph. Such subgraph can be regarded as a reliable input for the registration to ensure the accuracy and completeness of the reconstructed model. Snavely et al.\cite{snavely2008skeletal} computed a small skeletal subset of images based on the maximum leaf spanning tree. They reconstructed the skeletal set firstly and then added the remaining images via pose estimation. Yan et al.\cite{yan2017distinguishing} first introduced a geodesic consistency measure by selecting a set of iconic images. Correspondences that are connected in a visibility network but become disconnected according to visual propagation along the path network are geodetically inconsistent. Shen et al. \cite{shen2016graph} incrementally expanded the minimum spanning tree to form locally consistent strong triplets. 

The methods based on the analysis of the points or EGs aim to improve the quality of the view-graph. Due to the lack of higher contextual information, these methods usually remove a large number of positive edges, which tends to reduce the completeness of the reconstruction. In contrast, our method explores the spatial information among the regions in the scene to filter the correspondences from the ambiguous structures rather than directly remove the erroneous edges in the view-graph. This approach robustly detects ambiguous structures and improves the completeness of the reconstructed scene.

\subsubsection*{\bf Camera registration} Based on feature correspondences and EGs in the view-graph, registration is to determine the camera pose of each view. The structure of the scene is subsequently recovered based on camera poses.

The classical incremental pipeline, such as COLMAP \cite{schonberger2016structure}, typically adds an optimal image each time after an initial two-view reconstruction by repeatedly performing BA. This approach has the advantages of accuracy and robustness compared with the global approach \cite{sweeney2016large}. However, if a view is incorrectly registered due to lots of mismatched feature points, this error will be propagated to the other views of subsequent iterations, resulting in a completely incorrect scene structure. To overcome this challenge, Cui et al. \cite{cui2019efficient} proposed a batched incremental SfM framework that contains two iteration loops: the inner loop that selects a well-conditioned subset of tracks and the outer loop that uses rotations estimated via rotation averaging as weak supervision for the registration. Chen et al. \cite{chen2020graph} proposed a carefully designed clustering and merging algorithm to prevent the individual reconstructions from being affected by the wrong matches. Then, they performed a subgraph expansion step to enhance the connection and completeness of scenes. 

However, due to the influence of the erroneous correspondences, the SfM methods based on the optimization of registration usually struggle to recover the correct reconstruction while the scene structures share a strong visual resemblance. In our work, the ambiguous structures are detected by checking the pose consistency with the distinct structures, and the matches from the ambiguous structures are not involved in the registration. 

\subsubsection*{\bf Post-processing} For visually indistinguishable structures, some studies correct ambiguities based on a reconstructed 3D model with erroneous elements. Such method assumes that a priori knowledge of the ambiguous structure is not available at registration time, thereby resulting in reconstruction failure. Heinly et al. \cite{heinly2014correcting} proposed the informative measure of conflicting observations to identify the incorrectly placed unique scene structures. Later, Heinly et al. \cite{heinly2014recovering} presented another post-processing pipeline to split an incorrect reconstruction into error-free models by exploiting the co-occurrence information in the scene geometry with local clustering coefficients. As the post-processing methods take reconstructions as inputs, they work well on many challenging datasets due to more useful information about ambiguous structures. Obviously, complete reconstructions are indispensable for these methods, thereby introducing additional computation costs. 

Currently, most existing works focus on analyzing the EGs or points to handle the ambiguity, but do not well exploit the underlying high-level contextual information before reconstruction. Nevertheless, our method explores the spatial relationship among the regions of the scene based on track-communities to detect and correct ambiguous structures.

\section{Track-Community-Based SfM}
To correct the ambiguous structures, we propose a track-community-based SfM method (i.e., TC-SfM). Fig. \ref{fig_1} illustrates the pipeline of the proposed SfM method. This method first performs feature extraction and matching, geometric verification, and EG estimation to construct the view-graph as the conventional pipeline. To partition the scene according to the scene structure, a track-graph is constructed. Subsequently, the tracks are divided into groups by a community detection algorithm. Each track-community is regarded as a scene segment, which roughly refers to the local region of an object or several adjacent objects. Then, potential erroneous tracks are detected by diversity analysis on the track-graph. The segment that contains large erroneous tracks is potentially identified as ambiguous structures and corrected by checking the pose consistency with the help of other distinct segments. In this way, multiple ambiguity-free segments are obtained. Each segments is reconstructed via the standard incremental SfM. Finally, all partial reconstructions are aligned into a unified framework by a bidirectional consistency constraint. A final BA is performed to minimize the global reprojection error of the whole model. The TC-SfM method is described in detail in the following.

\subsection{Track-graph and Track-community Construction}
Unlike the existing methods that directly partition the scene based on the view-graph \cite{chen2020graph, bhowmick2014divide}, we exploit the dependencies among the tracks to explore the underlying scene structures by constructing the track-graph and track-community.

Firstly, a full view-graph is constructed. Based on the conclusion in the previous works \cite{yan2017distinguishing, cui2021view}, the image with the most matches to the given image is more likely to be the correct matched one. For a view pair $(C_i, C_j) $, we calculate the ratio of the number of the common 2D observations to the total observations in each view, which is denoted by $r^i_{ij}$ and $r^j_{ij}$, respectively. The weight of an edge in the view-graph is defined as the average of two ratios (i.e., $w_{ij} = \frac{r_{ij}^i + r_{ij}^j}{2} $).

Ideally, a track corresponds to a 3D point in the real world. If two 3D points are on the same object and are close to each other, their 2D projections in the view are usually closed accordingly. Therefore, the neighborhood relations of the tracks are utilized to explore the contextual information of the scene. For improving efficiency, the full tracks must be simplified. Inspired by the recent progress on superpixel structure in SLAM to improve the reconstruction \cite{concha2014using, wang2020relative}, we utilize superpixel segmentation to sample the tracks. We perform Simple Linear Iterative Clustering \cite{achanta2012slic} on each image to generate superpixels. If the superpixel region contains tracks, then the longest track will be selected because the long track is more reliable to represent the scene information in this superpixel. Note that the correspondences in EGs with weights less than $\tau_{w} $ are considered unreliable and ignored when searching correspondences for track sampling. While a track is sampled, the superpixels related to this track in other views will be skipped. All sampled tracks are regarded as nodes of the track-graph. If two tracks are visible in the same view and their 2D points are located in adjacent superpixels, then they are linked with an edge. The constructed track-graph can clearly display the surrounding information of a track.

\begin{figure}[!t]
	\centering
	\includegraphics[width=3.5in]{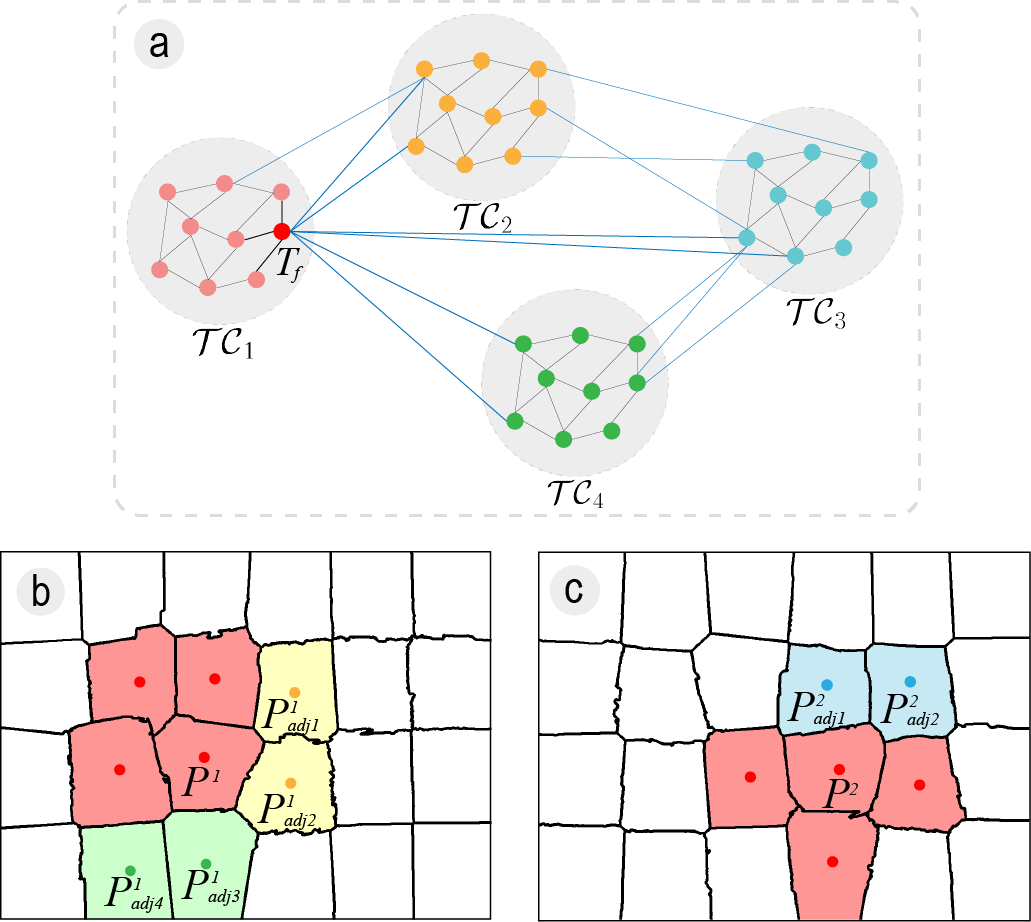}
	\caption{Illustration of the track-communities and the diversity analysis for a track. (a) Four detected track-communities. (b) and (c) Two keypoints in track $ T_f $ and their neighbors.  }
	\label{fig_2}
\end{figure}

Since the visual scenes are highly structured, the spatially proximal tracks exhibit strong dependencies, which carry high-level information about the structures of the scene as human perception. To exploit such information, the track-graph is initially split into $N$ track-communities inspired by a community detection algorithm in the network analysis\cite{Gu2015Image}. The community is composed of a set of tracks, which hold tight connections and correspond to the 2D keypoints on multiple views. 
Unlike the previous work \cite{shen2016graph, cui2017csfm} that explores the communities on the view-graph, each track-community in our method represents a local segment of the scene and typically belongs to an object or several adjacent objects in the scene. According to the 2D keypoints of the tracks on multiple views and the superpixels, the track-communities can be visualized as segments of the scene. In our implementation, the tracks are grouped by Louvain method \cite{blondel2008fast} for community detection. For example, Fig. \ref{fig_2}(a) shows the community detection result on the track-graph. The track-graph is divided into four communities $\{\mathcal{TC}_1, \mathcal{TC}_2, \mathcal{TC}_3, \mathcal{TC}_4\}$, representing four segments of the scene. Fig. \ref{fig_1}(c) shows the partitioning results in the sample images of \textit{Books} scene \cite{jiang2012seeing, roberts2011structure}, in which one community is labeled by one color.

\subsection{Ambiguous Structure Detection}
\begin{figure}[h!]
	\centering
	\includegraphics[width=3.5in]{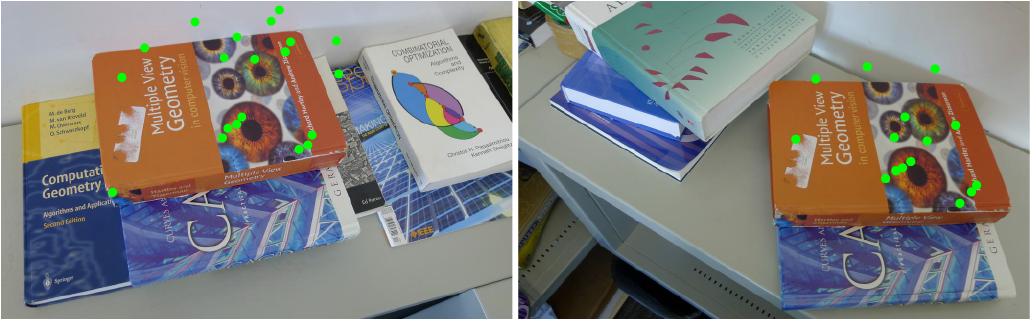}
	\caption{Potential erroneous tracks detected by GSI in two views of \textit{Books} scene.}
	\label{fig_3}
\end{figure}

This step aims to find potential ambiguous structures of the scene by exploiting the spatial contextual information provided by track-communities. The track that contains the mismatched correspondences is regarded as an erroneous track. Such track may cover different regions of the real world due to the local similar appearance. Intuitively, although such local regions have a similar appearance, their surrounding contents in the scene are usually different. Therefore, this challenge could be addressed by analyzing the surrounding information of a track in each view. The surrounding information of the correct track on each image is relatively consistent, while the erroneous track is more varied. Hence, we introduce Simpson's Diversity Index to measure the diversity of the surrounding track-communities. Simpson index is often used to quantify the biodiversity of habitat, which considers the number of species and the relative abundance of each species \cite{jost2006entropy}. The variant of Simpson index, called Gini-Simpson Index (GSI) is typically utilized to measure the diversity \cite{jost2006entropy}. In the track-graph, each track-community $ \mathcal{TC}_i$ is defined as species $\mathcal{S}_i $. If a track $T_f$ belongs to a species $\mathcal{S}_j $, then its adjacent tracks that belong to other species are regarded as individuals. We count the number $n_i $  of individuals of each species $\{\mathcal{S}_i,i \ne j\} $ in the adjacent tracks. Therefore, the GSI $gsi $ of a track can be calculated by:
\begin{equation}
	\label{eq_1}
	gsi = 1 - \sum_{i=1}^{N_{\mathcal{S}}}{(\frac{n_i}{N_{adj}})}^2,
\end{equation}
where $N_{adj} $ is the total number of the adjacent tracks that belong to other species; $N_{s} $ is the number of other species. GSI represents the diversity of the surrounding information of a track among different views. Although GSI cannot find all the erroneous tracks in the scene, such as the track located in the inner of the ambiguous structures, a large number of erroneous tracks can indicate that the community contains ambiguity. Accordingly, GSI is utilized to identify whether a community is ambiguous or distinct. For example, the dark red node $T_f$ in Fig. \ref{fig_2}(a) has three types of neighboring tracks that belong to other communities. Fig. \ref{fig_2}(b) and Fig. \ref{fig_2}(c) show two views that $T_f$ is visible. $P^1$  and $P^2$  are the corresponding 2D keypoints in the two views. $P^1$  has two adjacent tracks of $\mathcal{TC}_2$ and two adjacent tracks of $\mathcal{TC}_4$. $P^2$  has two adjacent tracks of $\mathcal{TC}_3$. A track with a large $gsi$ is regarded as a potential erroneous track. We set this threshold empirically as $\tau_{gs} = 0.5 $. Fig. \ref{fig_3} shows the potential erroneous tracks in the two views of \textit{Books} scene. The track-community where the ratio of erroneous tracks exceeds $\xi $  (in our work,  $\xi $ is set to 0.2) is regarded as ambiguous. That is, the corresponding segment contains different parts of the scene with a similar appearance. Otherwise, the segment is regarded as the distinct one.

\subsection{Ambiguous Structure Correction}
In this section, we introduce the method of correcting the ambiguous segments detected in the last step. During the incremental registration, 2D keypoints in the next candidate view and existing 3D points will be matched to calculate the camera pose. If the 2D-3D matches are all corrected, then the poses estimated by matches from different segments of the candidate view are consistent. However, the existence of ambiguous segments results in inconsistent poses. Based on this observation, we compare the difference between the pose calculated from the distinct segments and the pose calculated from the potential ambiguous segments. Then, the correspondences from the ambiguous segments are removed to correctly register the next view. After the correction, each segment is reconstructed.

\begin{figure}[t]
	\centering
	\includegraphics[width=3in]{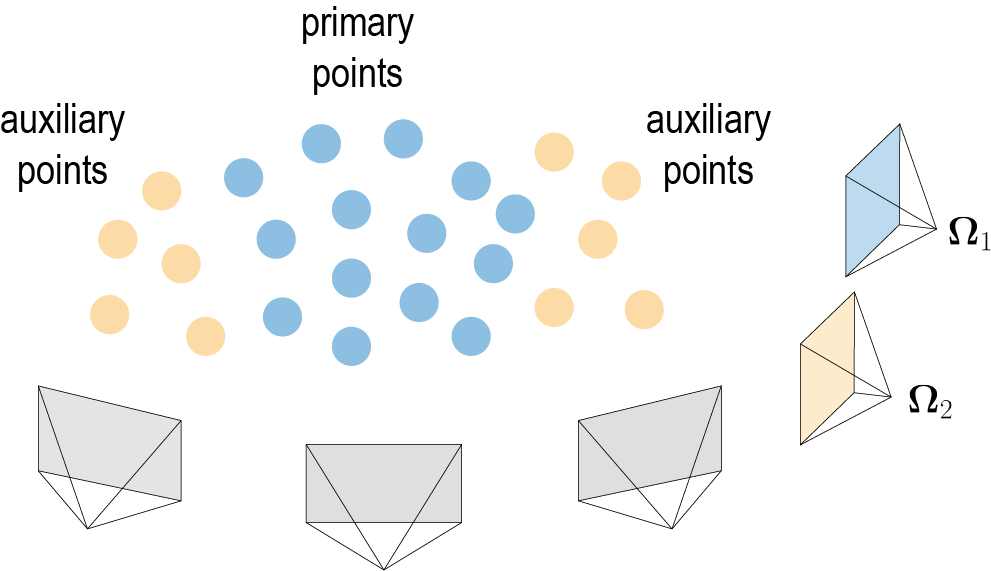}
	\caption{The pose of the next candidate view is estimated by 2D-3D matches from the primary and auxiliary points, respectively. If the consistency of $\boldsymbol{\Omega}_1 $ and $\boldsymbol{\Omega}_2 $ is not satisfied, then the matches that belong to the potential ambiguous segments are regarded as false.}
	\label{fig_4}
\end{figure}
For a given segment, the relevant views are collected first. If the tracks of this segment are visible in a view, then the view is related to this segment. Note that the images that only contain few tracks ($ < 30$) are ignored. Therefore, $N $  image clusters $\{ \mathbb{C}_1 , \mathbb{C}_2 , ..., \mathbb{C}_N\}$ are obtained, which correspond to $N $  segments, respectively. The correction is performed for the image clusters corresponding to the potential ambiguous segments individually to avoid erroneous correspondences contaminating the reconstruction. Thus, the sub-view-graph is constructed for each ambiguous image cluster by extracting nodes and edges from the full view-graph. For each sub-view-graph, the weights of all edges are sorted in a descending order. The two views associated with the edge of the largest weight are selected to produce an initial model via the initial two-view reconstruction. During the registration for each view of ambiguous image cluster, the triangulated 3D points that belong to the current segment are marked as \textit{primary points}, and the others that belong to distinct segments are marked as \textit{auxiliary points}. 

The image that matches the most primary points is selected as the next candidate view to incrementally reconstruct this segment and added to the reconstruction by registration and triangulation. When a new image is registered using the PnP algorithm \cite{kneip2011novel}, the keypoints of the new image that match the existing 3D points are triangulated. These matches are used to compute the relative pose and obtain the position and orientation of this new view. If this view fails to be registered due to large reprojection errors, then we will continue to try other images according to the number of matches previously mentioned. Let $\boldsymbol{\Omega}_1 $  be the pose of a successfully registered view. $\boldsymbol{\Omega}_1 $  is only estimated by using a 2D-3D match set $M_1 $  that is related to the primary points. However, $\boldsymbol{\Omega}_1 $  may be incorrect if this segment contains ambiguous structures. 

While an image contains a very similar appearance to the reconstructed scene, incorrectly matched 2D-3D correspondences may be preserved by RANSAC because they are relatively large. Accordingly, we ignore the keypoints that belong to the current segment and collect a 2D-3D match set $M_2 $  that corresponds to the auxiliary points. Then, another relative pose $\boldsymbol{\Omega}_2 $  is calculated by PnP with $M_2 $. Note that  $\boldsymbol{\Omega}_1 $ is calculated by only 2D-3D matches from the primary points, while $\boldsymbol{\Omega}_2 $  is calculated by only matches from the auxiliary points. In particular, $M_2$  belongs to the unambiguous segments of the scene. If the current image has no ambiguity, then these two poses should be consistent. Fig. \ref{fig_4} shows an illustration of this process. Consequently, we calculate the difference between $\boldsymbol{\Omega}_1 $  and  $\boldsymbol{\Omega}_2 $. The rotation error $e_r $  of the two poses could be expressed by  $acos(\frac{trace(\mathbf{R}_1^T\mathbf{R}_2)-1}{2}) $, where $\mathbf{R}_1 $  and $\mathbf{R}_2 $  are the rotation matrices from $\boldsymbol{\Omega}_1 $ and  $\boldsymbol{\Omega}_2 $, respectively. The translation error $e_t $ could be expressed by the angle error of two unit translation vectors, which is computed as $acos(\frac{\mathbf{t}_1}{\lVert \mathbf{t}_1 \rVert}\cdot \frac{\mathbf{t}_2}{\lVert \mathbf{t}_2 \rVert}) $, where $\mathbf{t}_1 $ and $\mathbf{t}_2 $ are the translation vectors from $\boldsymbol{\Omega}_1 $  and  $\boldsymbol{\Omega}_2 $, respectively. In our work, the pose of which the rotation error is more than $\epsilon_r(\epsilon_r=0.15) $  or the translation error is more than $\epsilon_t(\epsilon_t=0.35) $  will be regarded as inconsistent, and the current image will be rejected. In addition, the matches that are related to the current segment will be removed. Otherwise, the image is added to the reconstruction. After checking all images in the current image cluster $\mathbb{C}_i $, a consistent image subset will be obtained, which corresponds to an unambiguous segment. For the remaining images in $\mathbb{C}_i $, we repeat this process and obtain another consistent subset until there is no image in $\mathbb{C}_i $. The outputs of this process include one or several consistent segments and corresponding image clusters. 

When all ambiguous segments are corrected, each image cluster would be consistent and result in 3D models without ambiguity. Inspired by previous studies \cite{zhu2018very, chen2020graph} that take the partitioning strategy for SfM to overcome the drift problem, we also perform reconstruction for each segment individually in our method. After disambiguations, $N_1 $ consistent image clusters are obtained and correctly reconstructed by the traditional incremental SfM pipeline.  Note that the original correspondences will be cleaned by checking whether they lie in the same segment during registration. Due to the presence of many overlapping images between clusters, we do not need to reconstruct each cluster independently to prevent redundant registration. The image clusters are sorted by the number of images in a descending order. If the number of common images between two image clusters is larger than 20, then we merge the two image subsets. After registration, $N_2 $ local reconstructions are obtained.

\subsection{Local Reconstruction Merging}
This section aims to merge all the partial reconstructions into a complete 3D model in a unified framework. Each model is reconstructed in its local coordinate system originally. Two local models, $model_a $ and $model_b $, are merged by their relative similarity transformation  $\mathbf{T}_{ab}\in \mathbf{SIM}(3) $, including rotation transformation, translation transformation, and scale \cite{bhowmick2014divide}. If the two partial reconstructions can observe common 3D points, these 3D-3D correspondences can be used to fuse the 3D models by aligning two pieces of point clouds. Some existing methods try to find the overlapping views between two reconstructions \cite{chen2020graph}. However, the common view does not exist in many cases. Some other studies solve the transformation by following the image-to-image constraint across two clusters \cite{fang2019merge}. Nevertheless, the unreliable EGs, which cannot be filtered by geometric validation, limit their performance. Therefore, we propose a novel merging algorithm by taking the 3D-3D correspondences and two-view relative poses into account to accurately merge the models with bidirectional consistency. The initial similarity transformations are estimated by the relative poses of the camera pairs via three linear equations, and then they are all optimized by minimizing the reprojection error with bidirectional consistency. 

Let $\mathbf{T}_{ab}(\mathbf{R}_{ab}, \mathbf{t}_{ab}, s_{ab}) $ is the unknown relative similarity transformation from $model_a $ to $model_b $. $\mathbf{p}_k  $  is a 3D point that is visible in both two local reconstructions, and the local coordinates of $\mathbf{p}_k  $  are denoted as $\mathbf{p}_k^a  $ and $\mathbf{p}_k^b  $, respectively.  $C_i^a $ is a camera that can see $\mathbf{p}_k^a  $  in $model_a $ and  $C_j^b $ is a camera that can see $\mathbf{p}_k^b  $ in $model_b $. The pose of  $C_i^a $ in $model_a $can be denoted as $(\mathbf{R}_i^a, \mathbf{c}_i^a) $, where $\mathbf{R}_i^a $ is the rotation and $\mathbf{c}_i^a $ is the position. Similarly, $(\mathbf{R}_j^b, \mathbf{c}_j^b) $ is the pose of $C_j^b $ in   $model_b $. The relative transformation from  $C_i^a $ to $C_j^b $ can be denoted as $\mathbf{T}_{ij}^{ab}(\mathbf{R}_{ij}, \mathbf{t}_{ij}, s_{ab}, \lambda_{ij}^{ab}) $, where $\mathbf{t}_{ij} $  is a unit translation vector and $\lambda_{ij}^{ab} $  is the unknown scale.

\subsubsection*{\bf Relative rotation estimation}
We estimate the relative rotation between all partial reconstructions. The rotation between two 3D models can be obtained from EGs after cleaning the mismatches. The merged model should still satisfy the constraints between image pairs. Therefore, the relative rotation between the two 3D models can be estimated by using a linear equation system as:
\begin{equation}
	\label{eq_2}
	\mathbf{R}_j^{b}\mathbf{R}_{ab} = \mathbf{R}_{ij} \mathbf{R}_{i}^{a}.
\end{equation}

\subsubsection*{\bf Scale estimation}To calculate the scale between two reconstructions, the distance between 3D points transformed into the same frame is minimized. The scale factor can be estimated by:
\begin{equation}
	\label{eq_3}
	s_{ab} \mathbf{R}_{ij} (\mathbf{R}_i^a \mathbf{p}_k^a + \mathbf{t}_i^a) + \lambda_{ij}^{ab}\mathbf{t}_{ij} = \mathbf{R}_j^b\mathbf{p}_k^b + \mathbf{t}_j^b.
\end{equation}

\subsubsection*{\bf Relative translation estimation}With the rotation $\mathbf{R}_{ab} $  and scale $(s_{ab}, \lambda_{ij}^{ab}) $   fixed, the relative translation  $\mathbf{t_{ab}} $ between $model_a $  and $model_b $  are estimated based on the transformations of image pairs that cross two models. The linear equation system is defined as:
\begin{equation}
	\label{eq_4}
	\mathbf{t_{ab}} + s_{ab}\mathbf{R_{ab}}\mathbf{c}_i^a - \lambda_{ij}^{ab}(\mathbf{R}_j^b)^T\mathbf{t}_{ij} = \mathbf{c}_j^b.
\end{equation}

\begin{figure*}[!t]
	\centering
	\includegraphics[width=7.1in]{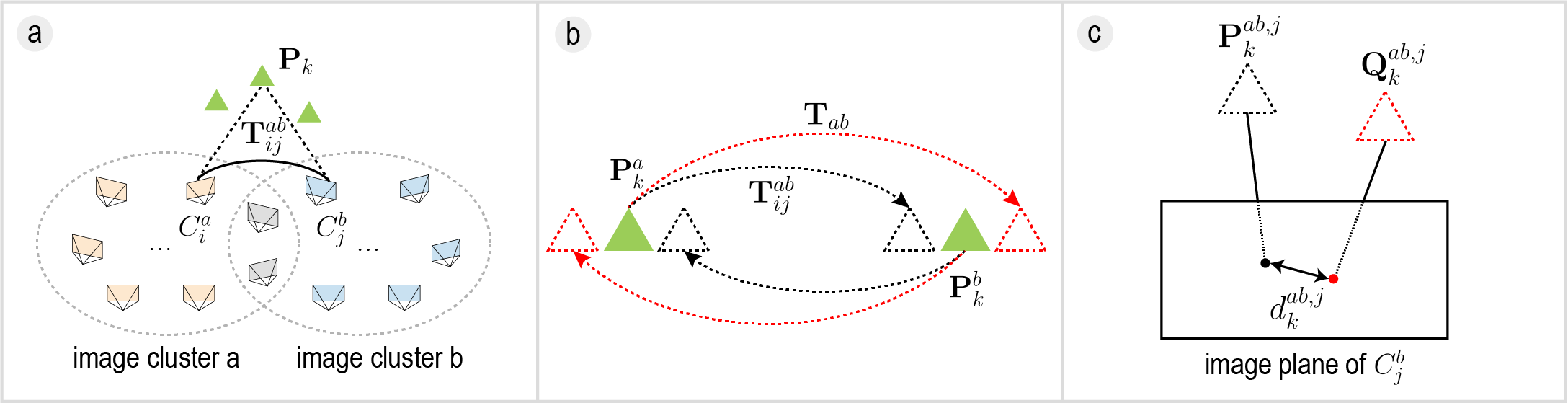}
	\caption{Illustration of bidirectional consistency cost. The 3D point   visible in two reconstructions are transformed from one model to another by two transformations. (a) Two partial reconstructions that share common 3D points. The green triangle denotes the common 3D point.  (b) Two types of transformations. The red and black dotted lines represent the 3D points transformed by the similarity transformation   and relative pose  , respectively. (c) Two transformed points are projected onto the image plane, and   is the distance between their projected points.}
	\label{fig_5}
\end{figure*}

\subsubsection*{\bf Bidirectional consistency optimization} After the initial values of similarity transformations between partial reconstructions are obtained, all the initial parameters will be further optimized. Here, a novel cost function is designed to accurately merge the models by enforcing the 3D-3D correspondence constraint and two-view relative rigid transformation constraint. As shown in Fig. \ref{fig_5}, the 3D point  $\mathbf{p}_k^a $ from the local coordinate system of $model_a $  can be transformed into the local coordinate system of $C_j^b $  in $model_b $  in two ways. We define $\mathbf{p}_k^{ab,j} $  as the 3D point transformed by the similarity transformation  $\mathbf{T}_{ab}(\mathbf{R}_{ab}, \mathbf{t}_{ab}, s_{ab}) $ between $model_a $   and $model_b $, and define  $\mathbf{q}_k^{ab,j} $ as the 3D point transformed by relative pose  $\mathbf{T}_{ij}^{ab}(\mathbf{R}_{ij}, \mathbf{t}_{ij}, s_{ab}, \lambda_{ij}^{ab}) $  between $C_i^a $  and $C_j^b $. Then, we have:
\begin{equation}
	\label{eq_5}
	\mathbf{p}_k^{ab,j} = \mathbf{R}_j^b(s_{ab}\mathbf{R}_{ab}\mathbf{p}_k^a + \mathbf{t}_{ab}) + \mathbf{t}_j^b,
\end{equation}
\begin{equation}
	\label{eq_6}
	\mathbf{q}_k^{ab,j} = s_{ab}\mathbf{R}_{ij}(\mathbf{R}_i^a \mathbf{p}_k^a+ \mathbf{t}_i^{a}) + \lambda_{ij}^{ab}\mathbf{t}_{ij}.
\end{equation}
We want to enforce the constraint that two transformations should be consistent with each other. Therefore, $\mathbf{p}_k^{ab,j} $  and  $\mathbf{q}_k^{ab,j} $ should be as close as possible. According to Eq. \ref{eq_3}, $\mathbf{q}_k^{ab,j} $is close to the point of transforming $\mathbf{p}_k^b $  to $C_j $. This indirectly enforces the 3D correspondences constraint. Furthermore, we utilize the reprojection error $d_k^{ab,j} $ to eliminate the range difference of the local models:
\begin{equation}
	\label{eq_6}
	d_k^{ab,j} = \left \|\mathcal{P}_j^b(\mathbf{p}_k^{ab,j}) - \mathcal{P}_j^b(\mathbf{q}_k^{ab,j}) \right \| ^2,
\end{equation}
where $\mathcal{P}_j^b(\mathbf{x}) $  means projecting the 3D point  $\mathbf{x} $ onto the image plane of $C_j $  in $model_b $. In the same way, $\mathbf{p}_k^b $  are also transformed by the inversed transformations. The distance between the projected points of them can be denoted as $d_k^{ba,i} $. Accordingly,  the bidirectional consistency cost function is formulated as:
\begin{equation}
	\label{eq_8}
	{E} = \sum\limits_{\substack{(a, b) \\
			a \ne b}}\sum\limits_{\substack{i\in \mathbb{C}_a \\
			j\in \mathbb{C}_b}}\sum\limits_{k\in\mathbb{P}_{ij}^{ab}}w_{ij}( d_k^{ab,j} + d_k^{ba,i} ),
\end{equation}
where  $w_{ij} $ is the edge weight of  $C_i $ and  $C_j $ in the view-graph defined in Section 3.1. We find the 3D point set  $\mathbb{P}_{ij}^{ab} $ observed by $C_i $  in  $model_a $ and $C_j $  in $model_b $  for all reconstruction pairs. The similarity transformations will be refined by minimizing the cost function ${E} $.

After all the pairwise similarity transformations are obtained, the partial reconstructions will be aligned to a global frame. Each partial reconstruction is regarded as the node, and two nodes will be connected if a similarity transformation exists between them. The weight of the edge is defined as the cost of Eq. \ref{eq_8}. The Minimum-cost Spanning Tree (MST) $\mathbb{T} $  of this graph is extracted to merge the models more accurately. $\mathbb{T} $ contains all $N_2 $  reconstructions and  $(N_2-1) $ pairwise transformations. Firstly, the edges that connect the leaf nodes are selected for merging. We merge the model with fewer images into the other via the refined similarity transformation. The MST $\mathbb{T} $  is updated by iteratively removing the leaf nodes. All the leaf nodes in  $\mathbb{T} $ and their neighbors are merged in the same way. At last, only one node is left in $\mathbb{T} $, and all the partial reconstructions are aligned into a unified frame. To make the 3D points and all camera poses more accurate, we perform the final BA on the merged reconstruction to minimize the reprojection error globally.

\section{Experiments and Discussions}
In this section, we evaluate the proposed TC-SfM from four types of datasets: ambiguous image datasets, sequential image datasets, unordered Internet image datasets, and our human body datasets. The specifications of these datasets are listed in Table \ref{tab_1}. The organization of the experiments is as follows:

\begin{table}[h]
	\begin{center}
		\caption{Specifications of the image datasets. The "GT" column reports whether the dataset has ground truth.}
		\begin{tabular}{@{}ccccc@{}}
			\toprule
			\begin{tabular}[c]{@{}c@{}}Dataset\\ type\end{tabular} & \# & \begin{tabular}[c]{@{}c@{}}\# of Images\\ per dataset\end{tabular} & GT  & \begin{tabular}[c]{@{}c@{}}Main evaluation\\ dimension\end{tabular}                   \\ \midrule
			Ambiguous                                              & 6  & 21–1559                                                            & No  & Disambiguation                                                                        \\
			Sequential                                             & 4  & 152-1108                                                           & No  & \multirow{2}{*}{Scalability \& Efﬁciency}                                             \\
			Unordered                                              & 15 & 247-5433                                                           & No  &                                                                                       \\
			Human body                                             & 8  & 90                                                                 & Yes & \begin{tabular}[c]{@{}c@{}}Quantitative performance \\ \& Ablation study\end{tabular} \\ \bottomrule
		\end{tabular}
		\label{tab_1}
	\end{center}

\end{table}

\begin{itemize}
	\item{Given that TC-SfM is targeted at the ambiguity problem, we firstly evaluate our method on six benchmark datasets, which are associated with visual ambiguities and common in the real world. Traditional SfM pipelines usually fail to recover correct scene structures on these datasets. Therefore, we compare our approach with recent representative methods for evaluating  disambiguation.}
	\item{In addition to focusing on ambiguity, we also refine the whole SfM pipeline. To evaluate its overall performance, we tested our method on the general image dataset, especially the sequential image datasets and the unordered Internet image datasets. These datasets are widely used in other SfM pipeline evaluations \cite{Ozyesil2015, Zhuang2018}. On these bases, we compare the performance of our method with state-of-the-art traditional as well as deep learning-based SfM methods. We also demonstrate the scalability of our method on some large-scale scenes included in the Internet image datasets.}
	\item{Finally, since the ground truth is difficult to obtain from existing datasets, we use our 3D human reconstruction system to capture eight human body image datasets where the camera pose ground truth is available. Quantitative evaluation is performed on human datasets to demonstrate the accuracy of our method in terms of rotation and position error. Moreover, we use human datasets to demonstrate the validity of the bidirectional consistency constraint in merging.}
\end{itemize} 

In our implementation, each segment of the scene is reconstructed based on the standard incremental pipeline of COLMAP \cite{schonberger2016structure} with the default configuration. Since the feature extraction and matching are common steps for SfM, the time consumption of these two steps is not included when we report the runtime of Tables \ref{tab_2} and \ref{tab_3}. The experiments were conducted on a PC equipped with an Intel Core i9-9900K CPU (3.60GHz), 128GB of RAM and an NVIDIA RTX 2080Ti GPU. Moreover, the configuration of the parameters and the limitations of our method are discussed.

\subsection{Evaluation on Ambiguous Image Datasets}
\begin{figure*}[!t]
	\centering
	\includegraphics[width=7in]{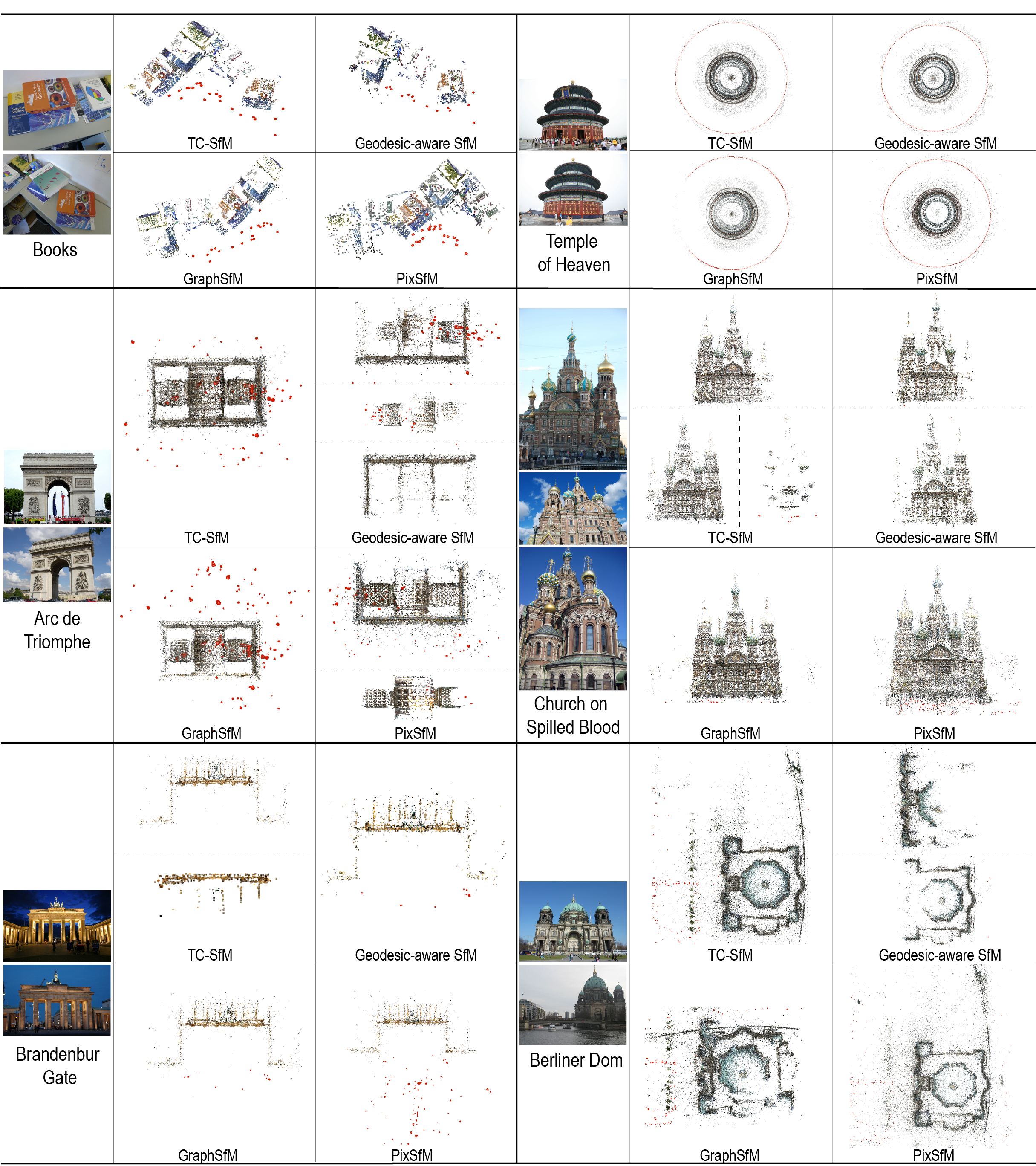}
	\caption{Comparison of the reconstruction results on the ambiguous image datasets by using the geodesic-aware SfM \cite{yan2017distinguishing}, PixSfM\cite{lindenberger2021pixel}, our TC-SfM and GraphSfM \cite{chen2020graph}, respectively. The first column shows the sampled images of the dataset. The separate models are split with dashed lines.}
	\label{fig_6}
\end{figure*}

We tested our TC-SfM on six ambiguous datasets, namely, \textit{Books} \cite{jiang2012seeing, roberts2011structure}, \textit{Temple of Heaven} \cite{shen2016graph}, \textit{Arc de Triomphe} \cite{heinly2014correcting}, \textit{Church on Spilled Blood} \cite{heinly2014correcting}, \textit{Brandenburg Gate} \cite{heinly2014correcting}, and \textit{Berliner Dom} \cite{heinly2014correcting}. \textit{Books} dataset has two same books placed in different locations. The content of the other datasets is all landmark architecture, which shares highly visual similarities on buildings. We compare the TC-SfM with two state-of-the-art methods for disambiguation, namely, geodesic-aware SfM proposed by Yan et al. \cite{yan2017distinguishing} and GraphSfM \cite{chen2020graph}. The geodesic-aware SfM \cite{yan2017distinguishing} is specifically designed for disambiguation, which is similar to our optimization goal. GraphSfM is a divide-and-conquer SfM method, but it is based on view-graph clustering as most of the existing methods. We also compare TC-SfM with a recent state-of-the-art learned SfM method, namely PixSfM \cite{lindenberger2021pixel}, based on the featuremetric refinement. The comparison results are shown in Fig. \ref{fig_6}. 

Benefiting from deep features and featuremetric optimization, PixSfM produces reasonable results on some ambiguous datasets like \textit{Berliner Dom}. However, for scenes with strong visual resemblance, such as \textit{Book}, \textit{Church on Spilled Blood} and \textit{Brandenburg Gate}, PixSfM incorrectly registers cameras and points of duplicated structures into the same place.

For GraphSfM, images are divided into clusters by view-graph clustering. The reconstruction performance greatly depends on graph cutting. However, the mismatched image pairs usually have large edge weights, causing them to be grouped into the same image cluster. Moreover, the merging of partial reconstructions also could be disturbed by erroneous correspondences. As shown in Fig. \ref{fig_6}, GraphSfM fails to obtain reasonable reconstructions on these ambiguous image datasets.

The geodesic-aware SfM\cite{yan2017distinguishing} performs well in sequential images with a uniform distribution and sufficient overlap, such as \textit{Temple of Heaven} dataset. However, the false EG removal and completeness of reconstruction are difficult to balance in the unordered Internet image datasets with various FoVs and illuminations. The EGs that do not satisfy particular conditions are directly rejected, which greatly affects completeness. For \textit{Arc de Triomphe} and \textit{Berliner Dom} datasets, the obtained several parts cannot be merged. In \textit{Book} and \textit{Brandenburg Gate} datasets, a part of the scene is missing in the result shown in Fig. \ref{fig_6}. If the thresholds of filtering are relaxed, then ambiguities could not be detected. 

In contrast, EGs are not directly handled in our TC-SfM. The scene is divided into several segments based on track-community to explore the underlying scene structures. The correspondences derived from different segments will be discarded during registration. Furthermore, erroneous correspondences belonging to ambiguous segments are removed by checking the pose consistent with distinct parts. For example, the scene of \textit{Book} dataset is divided into eight segments. The correspondences between the same two books are initially included in one segment. After ambiguity detection and correction, this segment splits into two unambiguous segments, and the images corresponding to these two books are divided into two clusters, resulting in correct partial 3D models. Meanwhile, the feature correspondences from the areas of the two books are discarded, thereby ensuring that the merging step is not misguided by false pairwise matches. For \textit{Arc de Triomphe}, \textit{Brandenburg Gate} and \textit{Church on Spilled Blood} datasets, the two sides of the building are divided into different segments. Thus, the scene is successfully recovered.

\begin{figure*}[!t]
	\centering
	\includegraphics[width=7in]{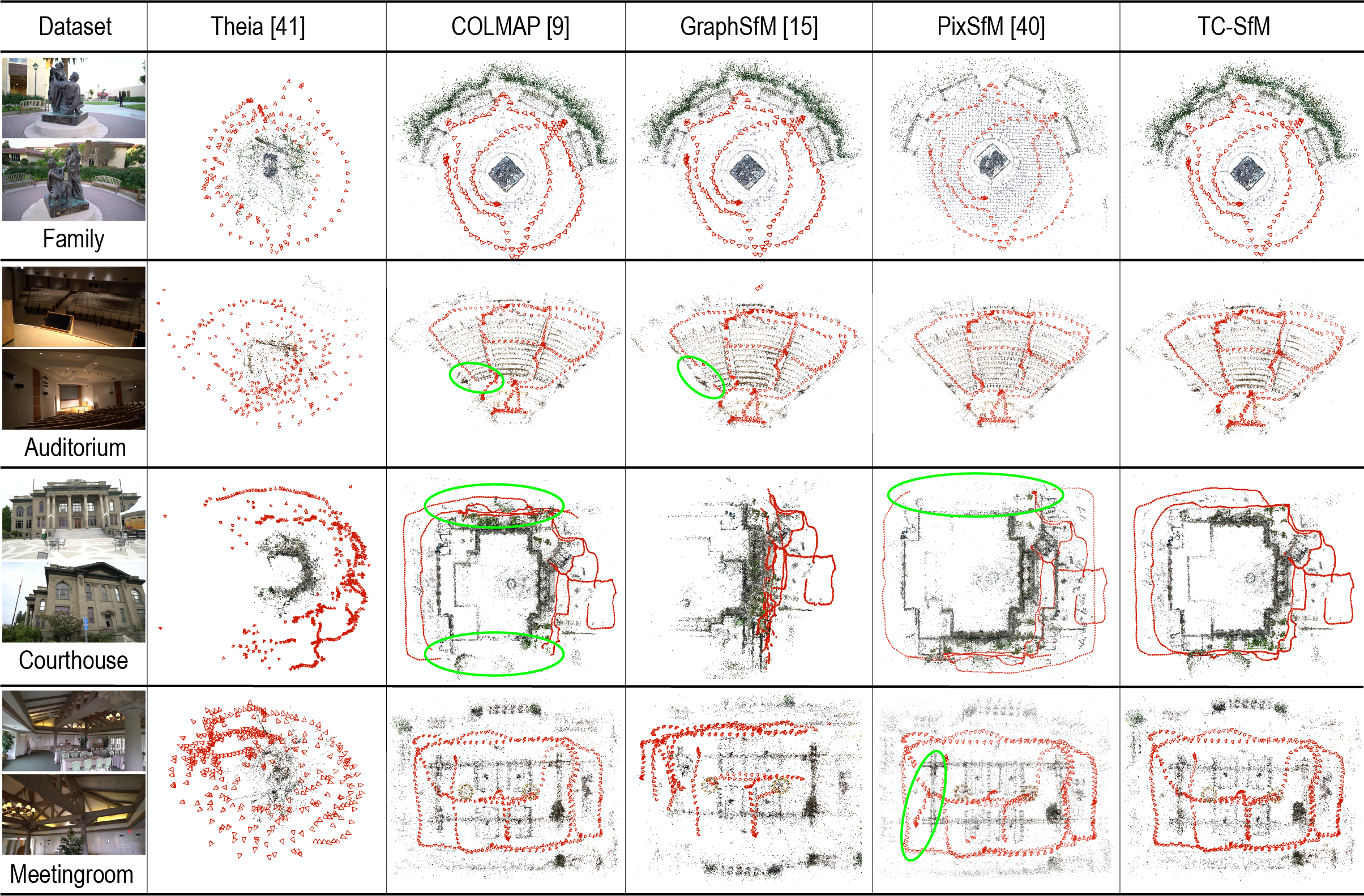}
	\caption{Reconstruction results of the four sequential image datasets. From left to right panels are reconstructed by Theia \cite{10.1145/2733373.2807405}, COLMAP \cite{schonberger2016structure}, GraphSfM \cite{chen2020graph}, PixSfM\cite{lindenberger2021pixel} and our TC-SfM, respectively. The camera poses are shown in red. The green ellipses mark the erroneous results of registration.}
	\label{fig_7}
\end{figure*}

\subsection{Evaluation on Sequential Image Datasets}

The proposed TC-SfM method is evaluated on the sequential image datasets from Tanks and Temples Dataset \cite{10.1145/3072959.3073599}. A uniformly distributed image set is provided for every scene from each high-resolution video sequence. Here, four representative image datasets, including outdoor and indoor environments, are selected for testing and comparison. Three state-of-the-art SfM methods, namely, incremental SfM (COLMAP \cite{schonberger2016structure}), a divide-and-conquer SfM (GraphSfM \cite{chen2020graph}), a global SfM (the global system of Theia library \cite{10.1145/2733373.2807405}) and a learning-based SfM (PixSfM \cite{lindenberger2021pixel}), are adopted and implemented for comparison. COLMAP and Theia are classic and widely used SfM systems. We use the default configurations in the implementation. GraphSfM and PixSfM are recent representative methods, which have advantages in terms of accuracy. The comparison results are shown in Fig. \ref{fig_7}.

In the outdoor dataset \textit{Family}, Theia failed to recover the structures, while the other four methods successfully reconstruct the scene. The other three datasets contain slight visual ambiguities. The indoor datasets \textit{Auditorium} and \textit{Meetingroom} exhibit repetitive furnishings. The outdoor dataset \textit{Courthouse} contains two same facades on the building. Although methods based on incremental SfM are more robust to correspondence outliers, COLMAP registers the ambiguous structures in the wrong location. GraphSfM is also disturbed by false matches in the merging step. PixSfM does not work well in the presence of large mismatches, such as in \textit{Courthouse} dataset. In contrast, TC-SfM achieves better results with respect to robustness due to the removal of matches from different segments. For example, in \textit{Auditorium} dataset, the seats in the three regions are divided into three segments by TC-SfM, thereby alleviating interference from the mismatches.

\subsection{Evaluation on Unordered Internet Image Datasets}
\begin{figure*}[!h]
	\centering
	\includegraphics[width=7.1in]{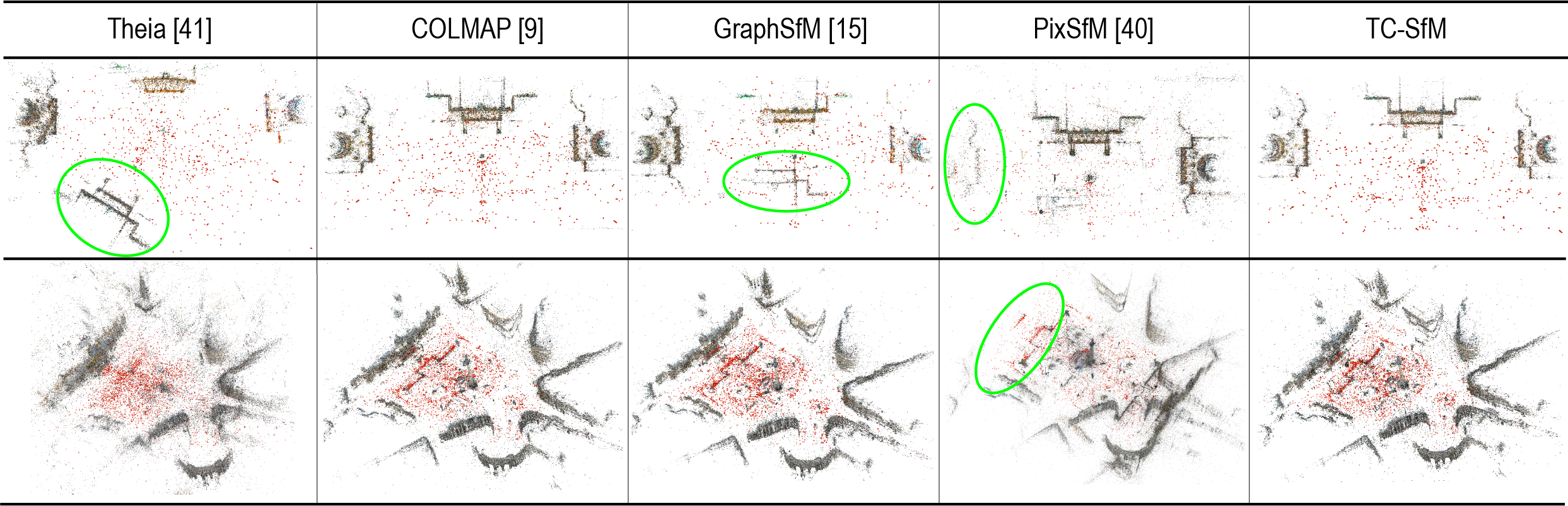}
	\caption{Reconstruction of the Internet image datasets for Theia \cite{10.1145/2733373.2807405}, COLMAP \cite{schonberger2016structure}, GraphSfM \cite{chen2020graph}, PixSfM\cite{lindenberger2021pixel} and our method. From top to bottom are \textit{GM} and \textit{Tr} datasets, respectively. The camera poses are shown in red. The green ellipses mark the erroneous results of registration.}
	\label{fig_8}
\end{figure*}

We evaluated TC-SfM on the unordered Internet image datasets from 1DSfM \cite{10.1007/978-3-319-10578-9_5}. Table \ref{tab_2} shows the comparison of the reconstruction results with COLMAP \cite{schonberger2016structure}, GraphSfM \cite{chen2020graph}, PixSfM \cite{lindenberger2021pixel} and the global system of Theia \cite{10.1145/2733373.2807405}, wherein the number of registered images, runtime and average reprojection error for each dataset are reported.  Theia and PixSfM generate the most registered images. However, these methods have large average reprojection errors and less accurate results (shown in Fig. \ref{fig_8}). Since the Internet image datasets are quite noisy, some views that have fewer correspondences and weak connections with others are missing in our method. Although the number of registered images by our method is less than other methods in several datasets, we can also recover similar structures with COLMAP. Fig. \ref{fig_8} shows the reconstruction results for these datasets. Note that for the dataset \textit{GM}, which contains ambiguous structures, other methods except for COLMAP produce incorrect models. The result shows that Theia and GraphSfM are sensitive to false correspondences, while our method achieves comparable performance with COLMAP in terms of robustness and accuracy. Theia and GraphSfM also show superiority in efficiency, while other methods are time-consuming. Benefiting from the partitioning scheme and mismatches correction, our method is faster than COLMAP for large-scale SfM tasks, such as the \textit{Tr} dataset in Table \ref{tab_2}.

\subsection{Application in Human Body Reconstruction}
\begin{figure}[h]
	\centering
	\includegraphics[width=3.5in]{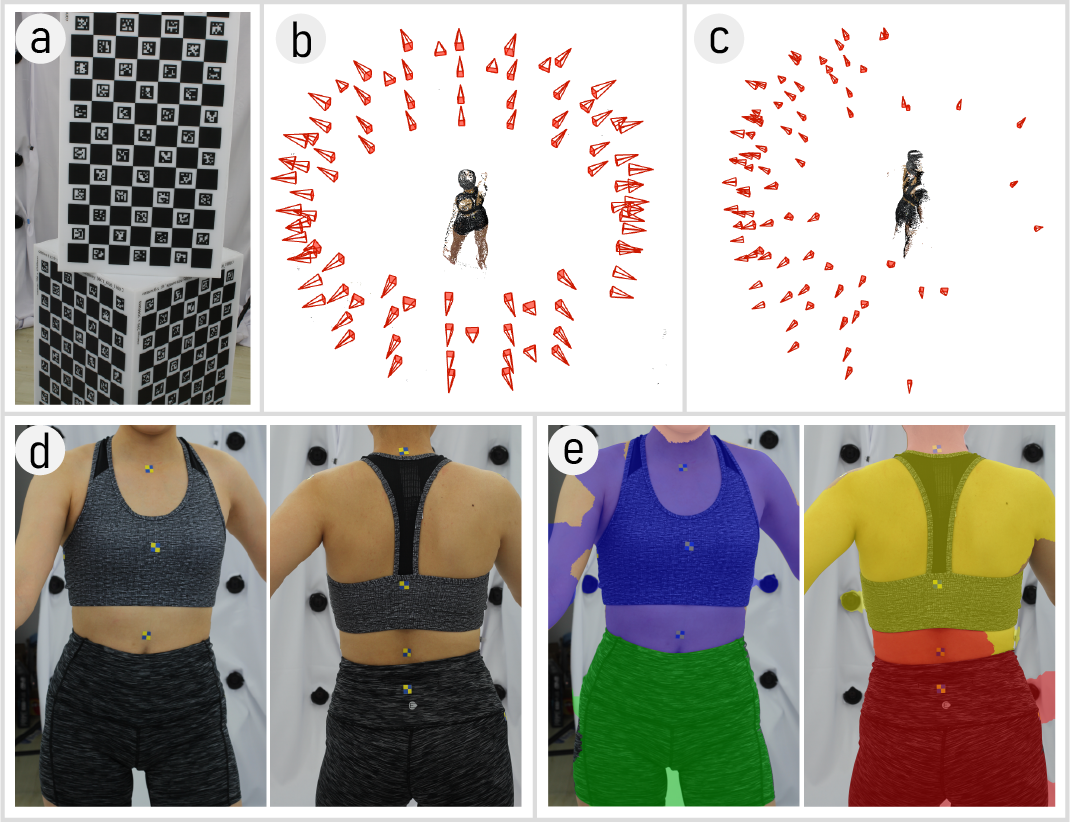}
	\caption{Illustration of the human body reconstruction. (a) Cylindrical calibration object of our acquisition system. (b) Camera layout and correctly reconstructed model. (c) Folded model affected by ambiguity. (d) Front and back of the human body, which share a highly similar appearance. (e) Structures of front and back disambiguated by our method.}
	\label{fig_9}
\end{figure}

We also apply our method to eight datasets with ground truth to quantitatively evaluate the accuracy compared with four state-of-the-art SfM approaches, namely, COLMAP \cite{schonberger2016structure}, GraphSfM \cite{chen2020graph}, the global system of Theia \cite{10.1145/2733373.2807405} and PixSfM \cite{lindenberger2021pixel}. Our human body acquisition system is equipped with 90 cameras and deployed in a cylindrical shape \cite{Zhang2021ARM}. The human body is located at the center of the cylinder. We utilize a cuboid calibration object (Fig. \ref{fig_9}(a)) with ChArUco patterns comparable to the human body size to calibrate all 90 cameras. The calibration results can serve as the ground truth due to the accurate feature extraction and matching of the calibration object. The quantitative results and runtime comparison are listed in Table \ref{tab_3} and \ref{tab_4}. 
\begin{table*}[!t]
	\begin{center}
		\caption{Comparison of the reconstruction performance on the unordered image dataset. $\#Reg $ denotes the number of registered images. $t$  denotes the runtime in seconds. $e$  denotes the average reprojection error. The datasets are listed in the first column. "Th", "CM", "GS", "PS" and "TS" are the abbreviation for Theia\cite{10.1145/2733373.2807405}, COLMAP\cite{schonberger2016structure}, GraphSfM\cite{chen2020graph}, PixSfM\cite{lindenberger2021pixel} and our TC-SfM, respectively.The best results are highlighted in bold font.}
		\label{tab_2}
		\begin{tabular}{@{}llllllllllllllllllll@{}}
			\toprule
			\multicolumn{1}{c}{\multirow{2}{*}{}} & \multicolumn{1}{c}{\multirow{2}{*}{$\#N$}} & \multicolumn{1}{c}{} & \multicolumn{5}{c}{$\#Reg$}                                                                                                    & \multicolumn{1}{c}{} & \multicolumn{5}{c}{$t$ [second] }                                                                                                      & \multicolumn{1}{c}{} & \multicolumn{5}{c}{$e$ [pixel]}                                                                                                      \\ \cmidrule(lr){4-8} \cmidrule(lr){10-14} \cmidrule(l){16-20} 
			\multicolumn{1}{c}{}                  & \multicolumn{1}{c}{}                     & \multicolumn{1}{c}{} & \multicolumn{1}{r}{Th} & \multicolumn{1}{r}{CM} & \multicolumn{1}{r}{GS} & \multicolumn{1}{r}{PS} & \multicolumn{1}{r}{TS} & \multicolumn{1}{c}{} & \multicolumn{1}{r}{Th} & \multicolumn{1}{r}{CM} & \multicolumn{1}{r}{GS} & \multicolumn{1}{r}{PS} & \multicolumn{1}{r}{TS} & \multicolumn{1}{r}{} & \multicolumn{1}{r}{Th} & \multicolumn{1}{r}{CM} & \multicolumn{1}{r}{GS} & \multicolumn{1}{r}{PS} & \multicolumn{1}{r}{TS} \\ \midrule
			Al&\multicolumn{1}{r}{627}&&\multicolumn{1}{r}{562}&\multicolumn{1}{r}{546}&\multicolumn{1}{r}{556}&\multicolumn{1}{r}{{\bf{568}}}&\multicolumn{1}{r}{543}&&\multicolumn{1}{r}{{\bf{659}}}&\multicolumn{1}{r}{2620}&\multicolumn{1}{r}{728}&\multicolumn{1}{r}{6386}&\multicolumn{1}{r}{2850}&&\multicolumn{1}{r}{1.39}&\multicolumn{1}{r}{{\bf{0.48}}}&\multicolumn{1}{r}{0.49}&\multicolumn{1}{r}{1.11}&\multicolumn{1}{r}{{\bf{0.48}}}\\
			EI&\multicolumn{1}{r}{247}&&\multicolumn{1}{r}{{\bf{231}}}&\multicolumn{1}{r}{228}&\multicolumn{1}{r}{229}&\multicolumn{1}{r}{218}&\multicolumn{1}{r}{230}&&\multicolumn{1}{r}{{\bf{35}}}&\multicolumn{1}{r}{365}&\multicolumn{1}{r}{263}&\multicolumn{1}{r}{965}&\multicolumn{1}{r}{636}&&\multicolumn{1}{r}{1.30}&\multicolumn{1}{r}{{\bf{0.74}}}&\multicolumn{1}{r}{0.75}&\multicolumn{1}{r}{1.19}&\multicolumn{1}{r}{{\bf{0.74}}}\\
			GM&\multicolumn{1}{r}{742}&&\multicolumn{1}{r}{{\bf{706}}}&\multicolumn{1}{r}{673}&\multicolumn{1}{r}{550}&\multicolumn{1}{r}{704}&\multicolumn{1}{r}{635}&&\multicolumn{1}{r}{{\bf{103}}}&\multicolumn{1}{r}{3133}&\multicolumn{1}{r}{759}&\multicolumn{1}{r}{7141}&\multicolumn{1}{r}{2789}&&\multicolumn{1}{r}{1.20}&\multicolumn{1}{r}{0.68}&\multicolumn{1}{r}{{\bf{0.67}}}&\multicolumn{1}{r}{1.13}&\multicolumn{1}{r}{{\bf{0.67}}}\\
			MM&\multicolumn{1}{r}{394}&&\multicolumn{1}{r}{{\bf{348}}}&\multicolumn{1}{r}{309}&\multicolumn{1}{r}{320}&\multicolumn{1}{r}{337}&\multicolumn{1}{r}{310}&&\multicolumn{1}{r}{{\bf{71}}}&\multicolumn{1}{r}{968}&\multicolumn{1}{r}{1265}&\multicolumn{1}{r}{2449}&\multicolumn{1}{r}{974}&&\multicolumn{1}{r}{0.96}&\multicolumn{1}{r}{0.50}&\multicolumn{1}{r}{0.50}&\multicolumn{1}{r}{1.14}&\multicolumn{1}{r}{{\bf{0.49}}}\\
			MN&\multicolumn{1}{r}{474}&&\multicolumn{1}{r}{{\bf{458}}}&\multicolumn{1}{r}{446}&\multicolumn{1}{r}{446}&\multicolumn{1}{r}{448}&\multicolumn{1}{r}{446}&&\multicolumn{1}{r}{{\bf{289}}}&\multicolumn{1}{r}{2120}&\multicolumn{1}{r}{1184}&\multicolumn{1}{r}{4082}&\multicolumn{1}{r}{2759}&&\multicolumn{1}{r}{1.25}&\multicolumn{1}{r}{{\bf{0.65}}}&\multicolumn{1}{r}{0.66}&\multicolumn{1}{r}{1.22}&\multicolumn{1}{r}{{\bf{0.65}}}\\
			ND&\multicolumn{1}{r}{553}&&\multicolumn{1}{r}{{\bf{545}}}&\multicolumn{1}{r}{532}&\multicolumn{1}{r}{536}&\multicolumn{1}{r}{518}&\multicolumn{1}{r}{536}&&\multicolumn{1}{r}{{\bf{412}}}&\multicolumn{1}{r}{8542}&\multicolumn{1}{r}{2371}&\multicolumn{1}{r}{6445}&\multicolumn{1}{r}{7855}&&\multicolumn{1}{r}{1.32}&\multicolumn{1}{r}{{\bf{0.64}}}&\multicolumn{1}{r}{0.69}&\multicolumn{1}{r}{1.15}&\multicolumn{1}{r}{{\bf{0.64}}}\\
			NL&\multicolumn{1}{r}{376}&&\multicolumn{1}{r}{339}&\multicolumn{1}{r}{320}&\multicolumn{1}{r}{320}&\multicolumn{1}{r}{{\bf{344}}}&\multicolumn{1}{r}{318}&&\multicolumn{1}{r}{{\bf{75}}}&\multicolumn{1}{r}{799}&\multicolumn{1}{r}{694}&\multicolumn{1}{r}{1809}&\multicolumn{1}{r}{1210}&&\multicolumn{1}{r}{1.40}&\multicolumn{1}{r}{{\bf{0.62}}}&\multicolumn{1}{r}{0.63}&\multicolumn{1}{r}{1.10}&\multicolumn{1}{r}{{\bf{0.62}}}\\
			PP&\multicolumn{1}{r}{354}&&\multicolumn{1}{r}{{\bf{339}}}&\multicolumn{1}{r}{320}&\multicolumn{1}{r}{325}&\multicolumn{1}{r}{338}&\multicolumn{1}{r}{321}&&\multicolumn{1}{r}{{\bf{60}}}&\multicolumn{1}{r}{678}&\multicolumn{1}{r}{341}&\multicolumn{1}{r}{1462}&\multicolumn{1}{r}{875}&&\multicolumn{1}{r}{1.42}&\multicolumn{1}{r}{{\bf{0.60}}}&\multicolumn{1}{r}{0.64}&\multicolumn{1}{r}{1.13}&\multicolumn{1}{r}{{\bf{0.60}}}\\
			Pic&\multicolumn{1}{r}{2508}&&\multicolumn{1}{r}{{\bf{2255}}}&\multicolumn{1}{r}{2133}&\multicolumn{1}{r}{2196}&\multicolumn{1}{r}{2180}&\multicolumn{1}{r}{2110}&&\multicolumn{1}{r}{{\bf{1130}}}&\multicolumn{1}{r}{17384}&\multicolumn{1}{r}{7453}&\multicolumn{1}{r}{67177}&\multicolumn{1}{r}{16530}&&\multicolumn{1}{r}{1.47}&\multicolumn{1}{r}{0.65}&\multicolumn{1}{r}{0.65}&\multicolumn{1}{r}{1.23}&\multicolumn{1}{r}{{\bf{0.64}}}\\
			RF&\multicolumn{1}{r}{1134}&&\multicolumn{1}{r}{{\bf{1079}}}&\multicolumn{1}{r}{1038}&\multicolumn{1}{r}{1029}&\multicolumn{1}{r}{1074}&\multicolumn{1}{r}{1030}&&\multicolumn{1}{r}{{\bf{295}}}&\multicolumn{1}{r}{6166}&\multicolumn{1}{r}{1923}&\multicolumn{1}{r}{9880}&\multicolumn{1}{r}{5042}&&\multicolumn{1}{r}{1.46}&\multicolumn{1}{r}{{\bf{0.59}}}&\multicolumn{1}{r}{0.67}&\multicolumn{1}{r}{1.21}&\multicolumn{1}{r}{{\bf{0.59}}}\\
			TL&\multicolumn{1}{r}{508}&&\multicolumn{1}{r}{{\bf{485}}}&\multicolumn{1}{r}{433}&\multicolumn{1}{r}{442}&\multicolumn{1}{r}{449}&\multicolumn{1}{r}{431}&&\multicolumn{1}{r}{{\bf{134}}}&\multicolumn{1}{r}{878}&\multicolumn{1}{r}{705}&\multicolumn{1}{r}{1378}&\multicolumn{1}{r}{1361}&&\multicolumn{1}{r}{1.21}&\multicolumn{1}{r}{{\bf{0.50}}}&\multicolumn{1}{r}{0.52}&\multicolumn{1}{r}{1.01}&\multicolumn{1}{r}{{\bf{0.50}}}\\
			Tr&\multicolumn{1}{r}{5433}&&\multicolumn{1}{r}{{\bf{4946}}}&\multicolumn{1}{r}{4744}&\multicolumn{1}{r}{4706}&\multicolumn{1}{r}{4856}&\multicolumn{1}{r}{4702}&&\multicolumn{1}{r}{{\bf{1369}}}&\multicolumn{1}{r}{51694}&\multicolumn{1}{r}{14764}&\multicolumn{1}{r}{148795}&\multicolumn{1}{r}{42398}&&\multicolumn{1}{r}{1.29}&\multicolumn{1}{r}{{\bf{0.61}}}&\multicolumn{1}{r}{0.64}&\multicolumn{1}{r}{1.19}&\multicolumn{1}{r}{{\bf{0.61}}}\\
			US&\multicolumn{1}{r}{930}&&\multicolumn{1}{r}{807}&\multicolumn{1}{r}{696}&\multicolumn{1}{r}{733}&\multicolumn{1}{r}{{\bf{841}}}&\multicolumn{1}{r}{730}&&\multicolumn{1}{r}{{\bf{46}}}&\multicolumn{1}{r}{3467}&\multicolumn{1}{r}{1258}&\multicolumn{1}{r}{4593}&\multicolumn{1}{r}{2588}&&\multicolumn{1}{r}{1.51}&\multicolumn{1}{r}{{\bf{0.62}}}&\multicolumn{1}{r}{0.68}&\multicolumn{1}{r}{1.12}&\multicolumn{1}{r}{{\bf{0.62}}}\\
			VC&\multicolumn{1}{r}{918}&&\multicolumn{1}{r}{{\bf{845}}}&\multicolumn{1}{r}{766}&\multicolumn{1}{r}{780}&\multicolumn{1}{r}{774}&\multicolumn{1}{r}{785}&&\multicolumn{1}{r}{{\bf{297}}}&\multicolumn{1}{r}{11105}&\multicolumn{1}{r}{3547}&\multicolumn{1}{r}{7645}&\multicolumn{1}{r}{9286}&&\multicolumn{1}{r}{1.39}&\multicolumn{1}{r}{{\bf{0.56}}}&\multicolumn{1}{r}{0.58}&\multicolumn{1}{r}{1.16}&\multicolumn{1}{r}{0.57}\\
			YM&\multicolumn{1}{r}{458}&&\multicolumn{1}{r}{428}&\multicolumn{1}{r}{411}&\multicolumn{1}{r}{415}&\multicolumn{1}{r}{{\bf{433}}}&\multicolumn{1}{r}{408}&&\multicolumn{1}{r}{{\bf{432}}}&\multicolumn{1}{r}{2123}&\multicolumn{1}{r}{1198}&\multicolumn{1}{r}{3056}&\multicolumn{1}{r}{2141}&&\multicolumn{1}{r}{1.32}&\multicolumn{1}{r}{{\bf{0.61}}}&\multicolumn{1}{r}{0.65}&\multicolumn{1}{r}{1.08}&\multicolumn{1}{r}{{\bf{0.61}}}\\
			\bottomrule
		\end{tabular}
	\end{center}
\end{table*}

\begin{table*}[t]
	\begin{center}
		\caption{Comparison of the number of registered images $\#Reg$ and runtime $t$ on eight human body datasets by using Theia \cite{10.1145/2733373.2807405}, COLMAP \cite{schonberger2016structure}, GraphSfM  \cite{chen2020graph}, PixSfM\cite{lindenberger2021pixel} and TC-SfM. The first column lists the datasets. The best results are highlighted in bold font.}
		\label{tab_3}
		\begin{tabular}{@{}llllllllllll@{}}
			\toprule
			\multicolumn{1}{c}{\multirow{2}{*}{}} & \multicolumn{5}{c}{$\#Reg$}                                                                                                                         & \multicolumn{1}{c}{} & \multicolumn{5}{c}{$t$ [second]}                                                                                                                           \\ \cmidrule(lr){2-6} \cmidrule(l){8-12} 
			\multicolumn{1}{c}{}                  & \multicolumn{1}{r}{Theia} & \multicolumn{1}{r}{COLMAP} & \multicolumn{1}{r}{GraphSfM} & \multicolumn{1}{r}{PixSfM} & \multicolumn{1}{r}{TC-SfM} & \multicolumn{1}{r}{} & \multicolumn{1}{r}{Theia} & \multicolumn{1}{r}{COLMAP} & \multicolumn{1}{r}{GraphSfM} & \multicolumn{1}{r}{PixSfM} & \multicolumn{1}{r}{TC-SfM} \\ \midrule

			\multicolumn{1}{c}{D1}&\multicolumn{1}{r}{{\bf{90}}}&\multicolumn{1}{r}{{\bf{90}}}&\multicolumn{1}{r}{88}&\multicolumn{1}{r}{37}&\multicolumn{1}{r}{{\bf{90}}}&&\multicolumn{1}{r}{54}&\multicolumn{1}{r}{190}&\multicolumn{1}{r}{{\bf{43}}}&\multicolumn{1}{r}{90}&\multicolumn{1}{r}{601}\\
			\multicolumn{1}{c}{D2}&\multicolumn{1}{r}{89}&\multicolumn{1}{r}{{\bf{90}}}&\multicolumn{1}{r}{89}&\multicolumn{1}{r}{27}&\multicolumn{1}{r}{{\bf{90}}}&&\multicolumn{1}{r}{51}&\multicolumn{1}{r}{188}&\multicolumn{1}{r}{{\bf{35}}}&\multicolumn{1}{r}{81}&\multicolumn{1}{r}{605}\\
			\multicolumn{1}{c}{D3}&\multicolumn{1}{r}{{\bf{90}}}&\multicolumn{1}{r}{{\bf{90}}}&\multicolumn{1}{r}{80}&\multicolumn{1}{r}{19}&\multicolumn{1}{r}{{\bf{90}}}&&\multicolumn{1}{r}{50}&\multicolumn{1}{r}{227}&\multicolumn{1}{r}{{\bf{21}}}&\multicolumn{1}{r}{70}&\multicolumn{1}{r}{571}\\
			\multicolumn{1}{c}{D4}&\multicolumn{1}{r}{{\bf{90}}}&\multicolumn{1}{r}{85}&\multicolumn{1}{r}{86}&\multicolumn{1}{r}{69}&\multicolumn{1}{r}{{\bf{90}}}&&\multicolumn{1}{r}{{\bf{81}}}&\multicolumn{1}{r}{1356}&\multicolumn{1}{r}{427}&\multicolumn{1}{r}{90}&\multicolumn{1}{r}{858}\\
			\multicolumn{1}{c}{D5}&\multicolumn{1}{r}{{\bf{90}}}&\multicolumn{1}{r}{86}&\multicolumn{1}{r}{89}&\multicolumn{1}{r}{57}&\multicolumn{1}{r}{{\bf{90}}}&&\multicolumn{1}{r}{{\bf{73}}}&\multicolumn{1}{r}{1549}&\multicolumn{1}{r}{99}&\multicolumn{1}{r}{96}&\multicolumn{1}{r}{857}\\
			\multicolumn{1}{c}{D6}&\multicolumn{1}{r}{{\bf{90}}}&\multicolumn{1}{r}{83}&\multicolumn{1}{r}{86}&\multicolumn{1}{r}{67}&\multicolumn{1}{r}{{\bf{90}}}&&\multicolumn{1}{r}{{\bf{74}}}&\multicolumn{1}{r}{1446}&\multicolumn{1}{r}{267}&\multicolumn{1}{r}{99}&\multicolumn{1}{r}{1032}\\
			\multicolumn{1}{c}{D7}&\multicolumn{1}{r}{89}&\multicolumn{1}{r}{81}&\multicolumn{1}{r}{72}&\multicolumn{1}{r}{66}&\multicolumn{1}{r}{{\bf{90}}}&&\multicolumn{1}{r}{108}&\multicolumn{1}{r}{1334}&\multicolumn{1}{r}{286}&\multicolumn{1}{r}{{\bf{97}}}&\multicolumn{1}{r}{992}\\
			\multicolumn{1}{c}{D8}&\multicolumn{1}{r}{{\bf{90}}}&\multicolumn{1}{r}{80}&\multicolumn{1}{r}{84}&\multicolumn{1}{r}{40}&\multicolumn{1}{r}{{\bf{90}}}&&\multicolumn{1}{r}{{\bf{73}}}&\multicolumn{1}{r}{1879}&\multicolumn{1}{r}{150}&\multicolumn{1}{r}{86}&\multicolumn{1}{r}{1552}\\
			\bottomrule
		\end{tabular}
	\end{center}
\end{table*}

\begin{table*}[!h]
	\begin{center}
		\caption{Comparison of rotation error $e_r $ and position error $e_c $ on eight human body datasets by using Theia \cite{10.1145/2733373.2807405}, COLMAP \cite{schonberger2016structure}, GraphSfM  \cite{chen2020graph}, PixSfM \cite{lindenberger2021pixel}, and TC-SfM with and without bidirectional consistency (BC) optimization. The datasets are listed in the first column. The best results are highlighted in bold font.  }
		\label{tab_4}
		\begin{tabular}{@{}llllllllllllll@{}}
			\toprule
			\multicolumn{1}{c}{\multirow{2}{*}{}} & \multicolumn{6}{c}{$e_r $ [degree]}                                                                                                                                                                                                       & \multicolumn{1}{c}{} & \multicolumn{6}{c}{$e_t$ [mm] }                                                                                                                                                                                                         \\ \cmidrule(lr){2-7} \cmidrule(l){9-14} 
			\multicolumn{1}{c}{}                  & \multicolumn{1}{r}{Theia} & \multicolumn{1}{r}{COLMAP} & \multicolumn{1}{r}{GraphSfM} & \multicolumn{1}{r}{PixSfM} & \multicolumn{1}{r}{\begin{tabular}[c]{@{}r@{}}TC-SfM\\ w/o BC\end{tabular}} & \multicolumn{1}{r}{TC-SfM} & \multicolumn{1}{r}{} & \multicolumn{1}{r}{Theia} & \multicolumn{1}{r}{COLMAP} & \multicolumn{1}{r}{GraphSfM} & \multicolumn{1}{r}{PixSfM} & \multicolumn{1}{r}{\begin{tabular}[c]{@{}r@{}}TC-SfM\\ w/o BC\end{tabular}} & \multicolumn{1}{r}{TC-SfM} \\ \midrule
			\multicolumn{1}{c}{D1}&\multicolumn{1}{r}{0.455}&\multicolumn{1}{r}{{\bf{0.023}}}&\multicolumn{1}{r}{0.045}&\multicolumn{1}{r}{36.302}&\multicolumn{1}{r}{{\bf{0.023}}}&\multicolumn{1}{r}{{\bf{0.023}}}&&\multicolumn{1}{r}{105.31}&\multicolumn{1}{r}{{\bf{2.08}}}&\multicolumn{1}{r}{5.67}&\multicolumn{1}{r}{18.950}&\multicolumn{1}{r}{2.11}&\multicolumn{1}{r}{2.10}\\
			\multicolumn{1}{c}{D2}&\multicolumn{1}{r}{0.482}&\multicolumn{1}{r}{{\bf{0.022}}}&\multicolumn{1}{r}{0.039}&\multicolumn{1}{r}{36.867}&\multicolumn{1}{r}{0.023}&\multicolumn{1}{r}{{\bf{0.022}}}&&\multicolumn{1}{r}{91.72}&\multicolumn{1}{r}{{\bf{2.03}}}&\multicolumn{1}{r}{4.50}&\multicolumn{1}{r}{27.759}&\multicolumn{1}{r}{2.10}&\multicolumn{1}{r}{2.07}\\
			\multicolumn{1}{c}{D3}&\multicolumn{1}{r}{0.440}&\multicolumn{1}{r}{{\bf{0.017}}}&\multicolumn{1}{r}{0.038}&\multicolumn{1}{r}{3.742}&\multicolumn{1}{r}{{\bf{0.017}}}&\multicolumn{1}{r}{{\bf{0.017}}}&&\multicolumn{1}{r}{103.20}&\multicolumn{1}{r}{2.38}&\multicolumn{1}{r}{4.86}&\multicolumn{1}{r}{21.661}&\multicolumn{1}{r}{2.29}&\multicolumn{1}{r}{{\bf{2.28}}}\\
			\multicolumn{1}{c}{D4}&\multicolumn{1}{r}{0.476}&\multicolumn{1}{r}{0.051}&\multicolumn{1}{r}{1.549}&\multicolumn{1}{r}{52.600}&\multicolumn{1}{r}{0.043}&\multicolumn{1}{r}{{\bf{0.042}}}&&\multicolumn{1}{r}{102.36}&\multicolumn{1}{r}{5.86}&\multicolumn{1}{r}{78.80}&\multicolumn{1}{r}{71.724}&\multicolumn{1}{r}{3.71}&\multicolumn{1}{r}{{\bf{3.60}}}\\
			\multicolumn{1}{c}{D5}&\multicolumn{1}{r}{0.497}&\multicolumn{1}{r}{0.080}&\multicolumn{1}{r}{0.092}&\multicolumn{1}{r}{48.515}&\multicolumn{1}{r}{0.070}&\multicolumn{1}{r}{{\bf{0.064}}}&&\multicolumn{1}{r}{103.69}&\multicolumn{1}{r}{7.00}&\multicolumn{1}{r}{10.07}&\multicolumn{1}{r}{28.652}&\multicolumn{1}{r}{6.18}&\multicolumn{1}{r}{{\bf{5.08}}}\\
			\multicolumn{1}{c}{D6}&\multicolumn{1}{r}{0.503}&\multicolumn{1}{r}{0.058}&\multicolumn{1}{r}{0.351}&\multicolumn{1}{r}{49.811}&\multicolumn{1}{r}{{\bf{0.056}}}&\multicolumn{1}{r}{{\bf{0.056}}}&&\multicolumn{1}{r}{104.21}&\multicolumn{1}{r}{6.54}&\multicolumn{1}{r}{20.36}&\multicolumn{1}{r}{29.735}&\multicolumn{1}{r}{7.47}&\multicolumn{1}{r}{{\bf{5.47}}}\\
			\multicolumn{1}{c}{D7}&\multicolumn{1}{r}{1.545}&\multicolumn{1}{r}{60.120}&\multicolumn{1}{r}{48.219}&\multicolumn{1}{r}{49.384}&\multicolumn{1}{r}{{\bf{0.045}}}&\multicolumn{1}{r}{{\bf{0.045}}}&&\multicolumn{1}{r}{182.65}&\multicolumn{1}{r}{1754.72}&\multicolumn{1}{r}{1607.80}&\multicolumn{1}{r}{61.632}&\multicolumn{1}{r}{{\bf{6.38}}}&\multicolumn{1}{r}{{\bf{6.38}}}\\
			\multicolumn{1}{c}{D8}&\multicolumn{1}{r}{0.441}&\multicolumn{1}{r}{41.495}&\multicolumn{1}{r}{36.642}&\multicolumn{1}{r}{70.839}&\multicolumn{1}{r}{0.044}&\multicolumn{1}{r}{{\bf{0.043}}}&&\multicolumn{1}{r}{104.22}&\multicolumn{1}{r}{1274.23}&\multicolumn{1}{r}{1233.26}&\multicolumn{1}{r}{50.727}&\multicolumn{1}{r}{5.88}&\multicolumn{1}{r}{{\bf{5.09}}}\\
			\midrule
		\end{tabular}	
	\end{center}
\end{table*}

Due to the similar structures of the clothes in the human body datasets  (Fig. \ref{fig_9}(d)), the original view-graph contains many incorrect EGs. Table \ref{tab_4} shows that Theia is disturbed by mismatches but obtains a relatively stable error in all human body datasets because the global approach evenly distributes residual errors. PixSfM obtains the worst performance for all human body datasets. It is not surprising that the deep learning-based method only recovers a few cameras and points in our human data, because such an approach is data-driven and might perform poorly in unfamiliar datasets. The human body datasets contain large low-texture skin and clothes regions, and these features might not be included in the training data of PixSfM. Therefore, PixSfM tends to reconstruct the structures with prominent texture (e.g., the silhouette of clothes), resulting in low completeness and accuracy. COLMAP and GraphSfM perform well on D1-D6, which have fewer false correspondences in the clothes. Although the local reconstruction of GraphSfM is based on COLMAP, it is still affected by outliers in the merging step and is worse than COLMAP. In the human datasets with challenging clothes (D7 and D8 in Table \ref{tab_4}), COLMAP and GraphSfM cannot distinguish between the front and back of the body because of the strong visual resemblance. The cameras located on the back are folded on the front side, as shown in Fig. \ref{fig_9}(c). Our method partitions the human body into several parts according to the track-communities. The front and back of the human body are regarded as two segments. The result of the partition is presented in Fig. \ref{fig_9}(e). The features belonging to different segments will not be matched to avoid the interference of false correspondences during registration and merging. The proposed TC-SfM can successfully recover all camera poses, while the ambiguity is seriously disruptive to the reconstruction result of COLMAP and GraphSfM. Thus, TC-SfM is more robust than others for ambiguities in the scene. In the human body datasets, TC-SfM achieves comparable or better results in terms of accuracy. 

PixSfM takes less time on human datasets due to a lot of missing cameras and points. Theia and GraphSfM are quite fast, while COLMAP and TC-SfM take more time for repetitive BA. Note that if the feature correspondence contains many outliers, COLMAP will spend more time reconstructing the best result. When the next candidate image fails to be registered, it will try another registration order, even a new initial pairwise reconstruction from scratch. Our method takes more time to construct the track-graph and detect ambiguity. However, the mismatched correspondences caused by ambiguous structures are removed before the registration, thereby reducing the number of retries in registration. 

To evaluate the performance of the bidirectional consistency cost, we test our human body datasets with and without bidirectional consistency optimization. The comparison of rotation and position error is reported in Table \ref{tab_4}. The optimization by bidirectional consistency further improves the accuracy of registration. For two local reconstructions, the similarity transformation between them can be estimated by common images or 3D correspondences. The minimization of the distances of the transformed posints is enforced. However, two image clusters do not always have common images, resulting in the merging failure. The 3D correspondences also contain outliers. Therefore, the pairwise EG constraint that crosses the two reconstructions is necessary for estimating the transformation.

\subsection{Parameter Configuration and Limitation}
Two key parameters  $\tau_{w} $ and  $\tau_{gs} $ should be manually set in our method, and the others can be set by default. $\tau_{w} $  is the edge weight threshold of the view-graph in the track sampling step, and $\tau_{gs} $  is the GSI threshold of the tracks in the track-graph. $\tau_{w} $  controls whether an image pair is rejected in the correspondence search. For the dataset with sufficient correspondence and stable illumination, such as the human body dataset, we set $\tau_{w} = 0.15 $  and $\tau_{gs} = 0.5 $. For the unordered Internet datasets that are considered noisy, $\tau_{w} = 0.05 $  and $\tau_{gs} = 0.65 $  are sufficient to produce satisfactory results. A large $\tau_{w} $  may result in broken models. A large $\tau_{gs} $ may result in insufficient detection of erroneous tracks.

Overall, our TC-SfM achieves superior performance on various datasets, even in the presence of ambiguous structures. We note that if a part of the scene is very weakly connected to other parts (i.e., with few matches with other views), then the track-community detection will consider this part as a separate one, which cannot be reconstructed individually due to the weak connections. The registration will ignore these views, resulting in the absence of a few views.

\section{Conclusion}
In this work, a track-community-based SfM method with a partitioning scheme is proposed to address the ambiguity caused by visually indistinguishable structures. The proposed track-community structure is used to partition the scene into several segments and introduce more contextual information, which makes it possible to detect potential ambiguous segments by analyzing the diversity of the neighborhood for each track. To distinguish the similar parts in the scene, we perform consistency validation between the poses estimated by the distinct and ambiguous segments. This approach enables a correct reconstruction because the erroneous correspondences are ignored during the registration. Then, each segment of the scene is individually reconstructed to mitigate the drift. The proposed bidirectional consistency cost can refine the pairwise similarity transformations between the local reconstructions, thereby further improving the merging accuracy. The experiments show that TC-SfM can effectively alleviate reconstruction failure resulting from ambiguity and achieve a more robust and accurate reconstruction. Given the absence of some weakly connected views, our future work is to employ the information between the different segments to further improve the reconstruction under extremely challenging cases.

\bibliographystyle{IEEEtran}
\bibliography{myref}

\newpage

\vfill

\end{document}